\documentclass{article}
\usepackage{ijcai25}

\usepackage{times}
\usepackage{soul}
\usepackage{url}
\usepackage[hidelinks]{hyperref}
\usepackage[utf8]{inputenc}
\usepackage[small]{caption}
\usepackage{graphicx}
\usepackage{amsmath}
\usepackage{amsthm}
\usepackage{booktabs}
\usepackage{algorithm}
\usepackage{algorithmic}
\usepackage{amsfonts}
\usepackage[switch]{lineno}

\title{Antibody Design and Optimization with Multi-scale Equivariant Graph Diffusion Models for Accurate Complex Antigen Binding}

\author{
Jiameng Chen$^1$
\and
Xiantao Cai$^{1}$\textsuperscript{*}\and
Jia Wu$^2$\And
Wenbin Hu$^{3,1}$\textsuperscript{*}\\
\affiliations
$^1$School of Computer Science, Wuhan University, Wuhan, China\\
$^2$Department of Computing, Macquarie University, Sydney, Australia\\
$^3$Wuhan University Shenzhen Research Institute, Shenzhen, China\\
\emails
\{jiameng.chen, caixiantao, hwb\}@whu.edu.cn,
jia.wu@mq.edu.au
}
\begin{document}

\maketitle
\let\thefootnote\relax
\footnotetext{\normalfont *Corresponding author}

\begin{abstract}
Antibody design remains a critical challenge in therapeutic and diagnostic development, particularly for complex antigens with diverse binding interfaces. Current computational methods face two main limitations: (1) capturing geometric features while preserving symmetries, and (2) generalizing novel antigen interfaces. Despite recent advancements, these methods often fail to accurately capture molecular interactions and maintain structural integrity. To address these challenges, we propose \textbf{AbMEGD}, an end-to-end framework integrating \textbf{M}ulti-scale \textbf{E}quivariant \textbf{G}raph \textbf{D}iffusion for antibody sequence and structure co-design. Leveraging advanced geometric deep learning, AbMEGD combines atomic-level geometric features with residue-level embeddings, capturing local atomic details and global sequence-structure interactions. Its E(3)-equivariant diffusion method ensures geometric precision, computational efficiency, and robust generalizability for complex antigens. Furthermore, experiments using the SAbDab database demonstrate a 10.13\% increase in amino acid recovery, 3.32\% rise in improvement percentage, and a 0.062~\AA\ reduction in root mean square deviation within the critical CDR-H3 region compared to DiffAb, a leading antibody design model. These results highlight AbMEGD's ability to balance structural integrity with improved functionality, establishing a new benchmark for sequence-structure co-design and affinity optimization. The code is available at: \url{https://github.com/Patrick221215/AbMEGD}.  

\end{abstract}

\section{Introduction}
Antibodies are essential for targeting disease-related antigens in the therapeutic and diagnostic fields \cite{carter2006potent}. Their binding specificity and affinity are governed by complementarity-determining regions (CDRs), which exhibit high sequence and structural diversity \cite{north2011new}. As a result, designing antibodies to effectively bind complex antigens requires addressing two key challenges: (i) simplifying geometric feature extraction while preserving critical symmetries, and (ii) enhancing model generalization to novel antigen interfaces. This is because current methods often fail to capture precise molecular interactions and preserve structural integrity.

\begin{figure}[t!]
    \centering
    \includegraphics[width=\linewidth]{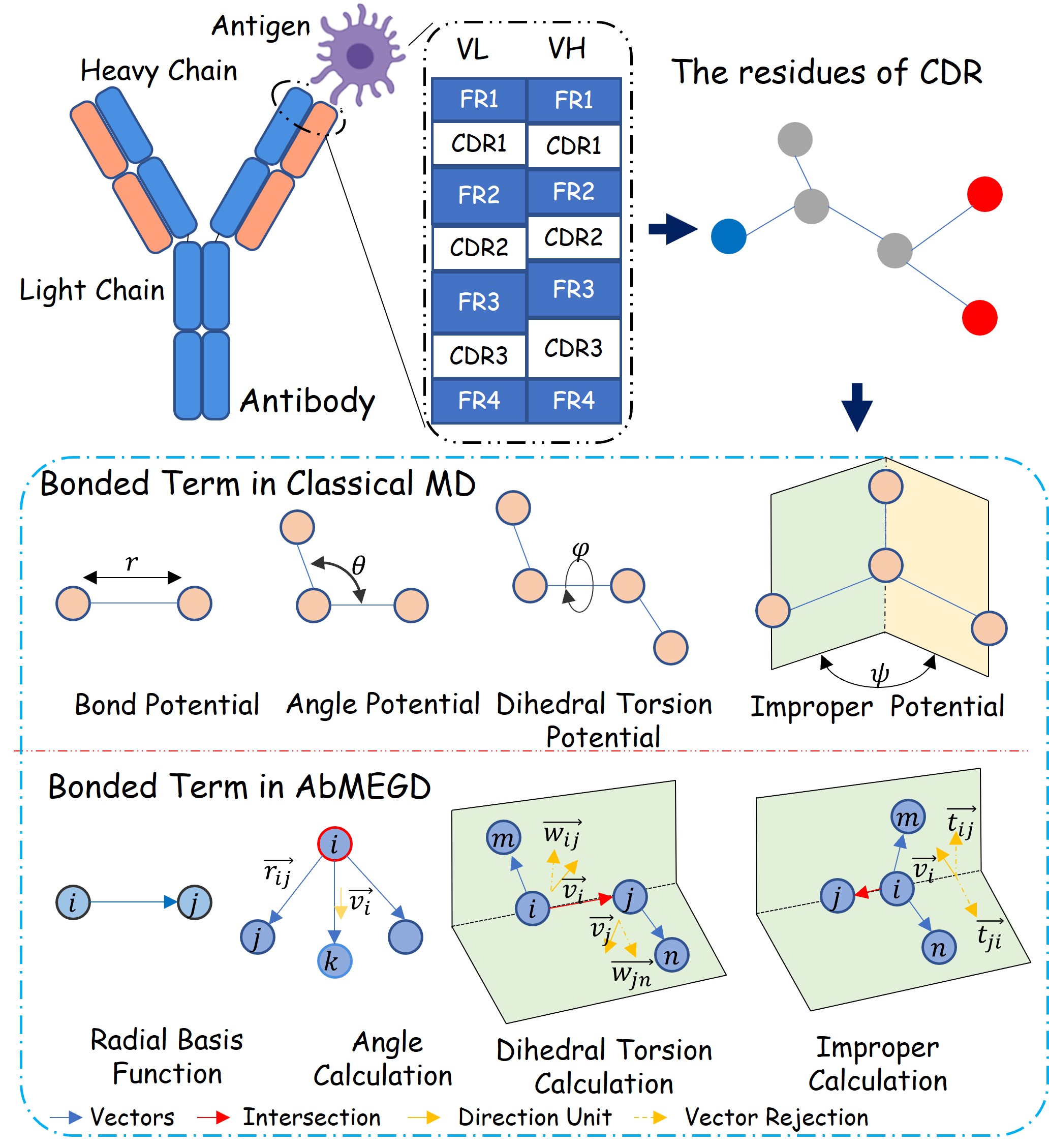} 
    \caption{The antibody-antigen complex structure and CDRs involved in AbMEGD's geometric calculations. This figure translates bonded terms from classical molecular dynamics (MD)—such as bond length, bond angle, dihedral torsion, and improper angle—into efficient linear-time model operations.}
    \label{fig:antiboyd_atom}
\end{figure}

First,the complexity of extracting geometric features while accurately modeling symmetries essential for antibody functionality poses a critical challenge in therapeutic antibody design. Thus, accurate CDR modeling is crucial, as geometric attributes such as bond lengths, angles, and torsions regulate antigen-binding capabilities \cite{carter2006potent}. However, maintaining symmetries such as rotation and translation invariance during feature extraction remains difficult. Traditional methods such as molecular dynamics (MD) and density functional theory provide precise geometric representations but are computationally restrictive in large-scale antibody modeling \cite{cai2024pretrainable,luo2022antigen}. Consequently, advances in geometric deep learning (GDL) have introduced symmetry-based inductive biases, enabling accurate molecular modeling with smaller datasets \cite{shen2023molecular}. For example, equivariant graph neural networks (EGNNs), central to GDL, efficiently capture molecular properties by embedding symmetries (e.g., rotation and translation invariance) \cite{garcia2021n}. In comparison, innovations such as GearBind improve binding affinity predictions through multi-relational graph construction and contrastive pretraining \cite{cai2024pretrainable}, while IgFold highlights the importance of structural precision in antibody design using advanced deep learning \cite{ruffolo2023fast}. Additionally, models like AbX integrate evolutionary and physical constraints with geometric factors to optimize antibody sequences \cite{zhu2024antibody}.

Second, generalizing antibody design to out-of-distribution antigen interfaces presents a significant challenge \cite{ijcai2024p234,ijcai2024p235}, complicating the development of effective antibodies. Antibody-antigen interactions are inherently unique and ad-hoc. This is because antibodies evolve rapidly to counter diverse pathogens without the long co-evolutionary histories typical of other protein-protein interactions \cite{adolf2018rosettaantibodydesign,kong2022conditional}. As a result, predictive models often fail to adapt to novel antigens that differ substantially from the existing training datasets \cite{joubbi2024antibody}. As antibody design continues to evolve, integrating advanced computational techniques with a deeper understanding of structural dynamics will be crucial for overcoming these challenges and unlocking the full potential of therapeutic antibodies.

To address these challenges, we propose AbMEGD, an end-to-end solution integrating Multi-scale Equivariant Graph Diffusion for co-designing antibody sequences and structures. Inspired by ViSNet \cite{wang2024enhancing}, which balances computational efficiency with effective geometric information utilization, AbMEGD incorporates geometry calculation and vector–scalar interactive equivariant message passing (ViS-MP). These strategies enable efficient modeling of atomic-level features such as bond angles, dihedral torsions, and improper angles, aligning with classical molecular dynamics force fields while ensuring geometric precision and scalability. AbMEGD combines atomic-level geometric features with residue-level embeddings in a multi-scale framework, capturing local atomic details and global sequence–structure interactions to maintain CDR structural fidelity. Leveraging ViS-MP, the model facilitates interactions between vector representations and scalar embeddings, fully exploiting geometric features while preserving E(3)-equivariance. This multi-scale architecture simplifies geometric feature extraction and enhances generalization to complex antigen interfaces. Unlike existing methods such as GEOAB \cite{pmlr-v235-lin24s}, which separate structure generation and affinity prediction, AbMEGD employs a unified framework that preserves essential geometric symmetries and addresses the computational challenges of high-throughput antibody design.

We validate AbMEGD using the SAbDab database \cite{dunbar2014sabdab}, which includes diverse antigen-antibody complexes. Compared to DiffAb, a leading model in antibody design, AbMEGD achieves a 10.13\% increase in amino acid recovery (AAR), a 3.32\% rise in improvement percentage (IMP), and a 0.062~\AA\ reduction in root mean square deviation (RMSD) within the critical CDR-H3 region. These results highlight its capacity to generate accurate structures and high-affinity antibodies, particularly for complex antigens. AbMEGD also excels in CDR design and affinity optimization tasks. Furthermore, its robust generalizability supports high-affinity antibody design for novel and out-of-distribution antigens, establishing a new benchmark in balancing structural conservation with improved functionality.

In summary, our contributions are as follows:
\begin{itemize}

\item \textbf{Unified Generative Framework with Multi-Scale Equivariant Graph Diffusion:} We propose AbMEGD, an end-to-end framework that unifies sequence-structure co-design and affinity optimization. By leveraging a multi-scale architecture, AbMEGD captures global sequence-structure interactions and local atomic details, preserving geometric symmetry while ensuring accuracy and efficiency.

\item \textbf{Superior Performance and Broad Applicability:} Evaluations on the SAbDab database confirm AbMEGD’s superior performance in sequence-structure co-design and affinity optimization. The model achieves structural fidelity and enhances functionality, showcasing adaptability across antibody design tasks and setting a new benchmark in therapeutic antibody development.

\end{itemize}

\begin{figure*}[ht]
    \centering
    \includegraphics[width=\textwidth]{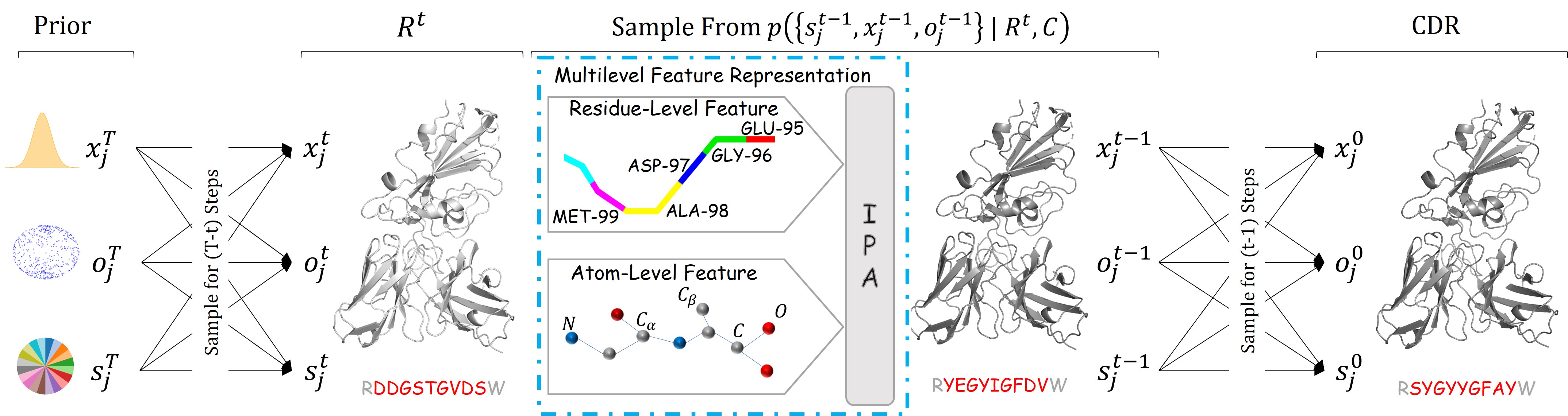} % 图片宽度设为全文宽度
    \caption{This figure depicts the generative diffusion process. At every iteration, the network uses the current state of the CDR as input to model the distribution for the sequences, positions, and orientations of the CDR in the subsequent step.}
    \label{fig:diffusion_progress}
\end{figure*}

\section{Related Work}
\subsection{Computational Antibody Design}

Traditional antibody design methods rely on statistical energy functions and Monte Carlo sampling for sequence-structure optimization \cite{adolf2018rosettaantibodydesign,ruffolo2021deciphering}, but suffer from inefficiency and local optima due to the rugged energy landscape. They also struggle with the vast sequence-structure space, especially for complex antigens, limiting design diversity. Deep learning has enabled sequence-based generative models that treat proteins as language \cite{alley2019unified,shin2021protein}, offering efficient sequence generation but often neglecting structural cues critical for antigen binding. The first CDR co-design model \cite{jin2021iterative}, targeting SARS-CoV-2, incorporated structure but depended on antigen-specific predictors, limiting generalizability. Moreover, it focused on backbone atoms while ignoring side-chain orientations essential for protein-protein interactions. In contrast, our method captures both backbone and side-chain features via multi-level representation, enabling a more complete modeling of the antigen-antibody interface for high-affinity binding.

\subsection{Diffusion-based Generative Model}
Diffusion probabilistic models (DDPMs) have gained traction for generating protein sequences and structures by denoising samples from a prior distribution \cite{sohl2015deep,song2019generative,ho2020denoising,martinkus2024abdiffuser}. Models like DiffAb \cite{luo2022antigen} and DIFFFORCE \cite{kulyte2024improving} have demonstrated superior efficiency in generating antibody sequences and structures. However, these models primarily focus on residue-level features and often fail to capture atomic-level interactions essential for high-specificity antibody-antigen binding. In comparison, our approach enhances existing DDPMs using VisNet-inspired graph-based modules \cite{wang2024enhancing} to model atomic-level geometric features. This facilitates the effective representation of complex three-dimensional (3D) spatial relationships among the atoms in antibodies and antigens, optimizing antibody specificity and affinity through detailed atomic interaction modeling.

\section{Methodology}
This section is organized as follows: Section~\ref{sec:notations} introduces the notations used throughout this paper and formally defines the problem. Section~\ref{sec:Multilevel} presents the multi-level feature generation and unified representation mechanism, integrating residue- and atom-level features. Finally, Section~\ref{sec:diffusion} details the diffusion processes tailored for modeling antibodies. 
% Finally, Section~\ref{sec:sampling} describes the sampling algorithms designed for various antibody design tasks.

\subsection{Problem Definitions and Notations}
\label{sec:notations}
In this study, we represent each amino acid in a protein complex using three components: type, $C_\alpha$ atom coordinates, and orientation. Specifically, the amino acid type is denoted as \( s_{r_i} \in \{ACDEFGHIKLMNPQRSTVWY\} \), where \( i = 1, \dots, N \) and \( N \) is the total number of amino acids in the complex. The spatial position of each residue is described by the $C_\alpha$ coordinates \( \mathbf{x}_{r_i} \in \mathbb{R}^3 \), and the orientation is denoted as \( \mathbf{O}_{r_i} \in SO(3) \), which captures the geometric directionality in 3D space.

We consider the antigen structure and the antibody framework as given, focusing our design efforts on the CDRs of the antibody. Specifically, let the generated CDR consist of \( m \) amino acids, indexed from \( c+1 \) to \( c+m \), the set of residues in the CDR is denoted as \( \mathcal{R} = \{(s_{r_j}, \mathbf{x}_{r_j}, \mathbf{O}_{r_j}) \mid j = c+1, \dots, c+m\} \). Our objective is to model the conditional distribution \( P(\mathcal{R} \mid \mathcal{C}) \), where \( \mathcal{C} = \{(s_{r_i}, \mathbf{x}_{r_i}, \mathbf{O}_{r_i}) \mid r_i \in \{1, \dots, N\} \setminus \{c+1, \dots, c+m\}\} \) represents the antigen-antibody framework excluding the CDR regions.

\subsection{Multi-level Feature Generation and Unified Representation Mechanism}
\label{sec:Multilevel}

The overall workflow of AbMEGD is illustrated in Figure~\ref{fig:diffusion_progress}. This study presents an innovative multi-level feature generation and unified representation mechanism that seamlessly integrates residue- and atom-level features. This approach leverages the complementary nature of these features to simultaneously capture global structural information and fine-grained geometric details, enabling comprehensive characterization of antibody-antigen complexes.

\paragraph{Residue-Level Feature Generation.} At the residue level, we employ a multi-layer perceptron (MLP) module to process features of individual residues and residue pairs. We define the following variables to represent the feature states of the CDRs:$\{ s_{r_j}^{t}, \mathbf{x}_{r_j}^{t}, \mathbf{O}_{r_j}^{t} \}_{j=1}^{c+m} \cup \{ s_{r_i}^{t}, \mathbf{x}_{r_i}^t, \mathbf{O}_{r_i}^{t} \}_{i \in \{1, \ldots, N\} \setminus \{c+1, \ldots, c+m\}},$ where \( s_{r_j}^{t} \) denotes the amino acid type of residue \( j \) at time step \( t \), \( \mathbf{x}_{r_j}^{t} \) represents the C\(_\alpha\) backbone coordinates, and \( \mathbf{O}_{r_j}^{t} \in SO(3) \) indicates the orientation.

The single-residue features encapsulate each residue's chemical properties, geometric characteristics, and local structural environment. Thus, the features for residue \( r_i \) are defined as~\cite{luo2022antigen}:
\begin{equation}
    \mathbf{e}_{r_i} = \text{MLP}_{\text{res}}([\mathbf{t}_{r_i}, \mathbf{g}_{r_i}, \mathbf{d}_{r_i}, \mathbf{f}_{r_i}]),
    \label{eq:single_residue_features}
\end{equation}
where \( \mathbf{t}_{r_i} \) represents the amino acid type features, obtained by embedding the amino acid type \( s_{r_i} \) into a low-dimensional vector. \( \mathbf{g}_{r_i} \) denotes the local geometric features, derived by constructing a local reference frame based on the backbone atomic coordinates \( \mathbf{p}_{r_i} \). \( \mathbf{d}_{r_i} \) encodes the backbone dihedral angles through angular encoding of \( \phi_{r_i} \) and \( \psi_{r_i} \), which describe the torsional conformation of the protein backbone. Finally, \( \mathbf{f}_{r_i} \) represents the fragment type features, capturing local fragment-level structural information.

Pairwise residue features describe the interactions and geometric relationships between residues. The features for a residue pair \((r_i, r_j)\) are represented by ~\cite{luo2022antigen}:
\begin{equation}
    \mathbf{z}_{r_i,r_j} = \text{MLP}_{\text{pair}} \left( [\mathbf{t}_{r_i,r_j}, \mathbf{r}_{r_i,r_j}, \mathbf{d}_{r_i,r_j}, \mathbf{o}_{r_i,r_j}] \right),
    \label{eq:pairwise_residue_features}
\end{equation}
where \( \mathbf{t}_{r_i,r_j} \) represents the residue-pair type features, encoded by embedding the combination of the amino acid types \( s_{r_i} \) and \( s_{r_j} \). \( \mathbf{r}_{r_i,r_j} \) indicates the relative positional features of the residue pair, derived from the sequence index difference \( \Delta_{r_i,r_j} = |r_i - r_j| \). \( \mathbf{d}_{r_i,r_j} \) denotes the geometric distance features, calculated as the Euclidean distance between the backbone atoms of the two residues \( \| \mathbf{p}_{r_i} - \mathbf{p}_{r_j} \| \) and further represented using Gaussian distribution encoding. Lastly, \( \mathbf{o}_{r_i,r_j} \) refers to the orientation features of the residue pair, based on the relative orientation angles (\( \phi_{r_i,r_j} \), \( \psi_{r_i,r_j} \)). Specifically, \( \phi_{r_i,r_j} \) and \( \psi_{r_i,r_j} \) capture spatial orientation and torsional relationships, ensuring rotationally invariant geometric properties.

\paragraph{Atom-Level Feature Generation.} To enhance complex representation, we focus on heavy atoms (\( \text{N}, \text{C}_\alpha, \text{C}, \text{O}, \) and side-chain \( \text{C}_\beta \), excluding glycine which lacks \( \text{C}_\beta \)). The atom-level feature extraction process utilizes the positions of selected atoms \( \mathbf{X} \in \mathbb{R}^{A \times 3} \) and their atomic indices \( \mathbf{Z} \in \mathbb{N}^{A} \), where \( A \) is the number of heavy atoms.

Initially, an embedding block encodes the atomic indices and pairwise edge distances into a high-dimensional embedding space. Then, a series of ViS-MP blocks iteratively update the heavy atoms' scalar and vector representations based on their interactions. Subsequently, residual connections are incorporated between consecutive ViS-MP blocks \cite{wang2024enhancing}. Thus, the final output includes node and edge features, representing the molecular structure at the atomic level.

Furthermore, the embedding process starts with the heavy atomic indices \( z_{a_i} \) of atom \( a_i \), embedded along with the pairwise distances \( e_{a_ia_j} \) using radial basis functions within a cutoff distance. The initial node and edge embeddings are defined as~\cite{wang2024enhancing}:
\begin{equation}
\mathbf{h}_{a_i}^0, \mathbf{f}_{a_ia_j}^0 = \text{Embedding}(\mathbf{z}_{a_i}, \mathbf{z}_{a_j}, \mathbf{e}_{a_ia_j}), \quad a_j \in \mathcal{N}(a_i),
\label{eq:atom_embedding}
\end{equation}
where \( \mathcal{N}(a_i) \) denotes the set of first-order neighbors of atom \( a_i \), and \( a_j \) represents one of its neighbors. The initial vector embedding \( \mathbf{v}_{a_i}^0 \) is initialized to \( \mathbf{0} \) and projected into the embedding space as \( \mathbf{v}_{a_i}^0 \in \mathbb{R}^{3 \times K} \), where \( K \) is the hidden dimension size.

% \textbf{Vector-Scalar Message Passing} 
Moreover, the ViS-MP mechanism facilitates the interaction between scalar and vector features, enhancing the geometric representation of the molecular structure. Therefore, the mechanism's key operations expressed as follows \cite{wang2024enhancing}:
\begin{align}
    m_{a_i}^l &= \sum_{a_j \in \mathcal{N}(a_i)} \phi_m^s \left( h_{a_i}^l, h_{a_j}^l, f_{a_i,a_j}^l \right), \label{eq:message_scalar} \\
    \mathbf{m}_{a_i}^l &= \sum_{a_j \in \mathcal{N}(a_i)} \phi_m^v \left( m_{a_i,a_j}^l,\mathbf{r}_{a_i,a_j}, \mathbf{v}_{a_j}^l \right), \label{eq:message_vector} \\
    h_{a_i}^{l+1} &= \phi_{\text{un}}^s \left( h_{a_i}^l, m_{a_i}^l, \langle \mathbf{v}_{a_i}^l, \mathbf{v}_{a_i}^l \rangle \right), \label{eq:update_scalar} \\
    f_{a_i,a_j}^{l+1} &= \phi_{\text{ue}}^s \left( f_{a_i,a_j}^l, \langle \text{Rej}_{\mathbf{r}_{a_i,a_j}} (\mathbf{v}_{a_i}^l), \text{Rej}_{\mathbf{r}_{a_j,a_i}} (\mathbf{v}_{a_j}^l) \rangle \right), \label{eq:update_edge} \\
    \mathbf{v}_{a_i}^{l+1} &= \phi_{\text{un}}^v \left( \mathbf{v}_{a_i}^l, m_{a_i}^l, \mathbf{m}_{a_i}^l \right), \label{eq:update_vector}
\end{align}
Here, \( \mathbf{h}_{a_i}^l \) denotes the scalar embedding of atom \( a_i \) at layer \( l \), \( f_{a_i,a_j}^l \) stands for the edge feature between atoms \( a_i \) and \( a_j \) at layer \( l \), and \( \mathbf{v}_{a_i}^l \) represents the vector embedding of atom \( a_i \) at layer \( l \). Meanwhile, \( m_{a_i}^l \) is the aggregated scalar message for atom \( a_i \) at layer \( l \), computed by the scalar message function \( \phi_m^s \). Similarly, \( \mathbf{m}_{a_i}^l \) is the aggregated vector message for atom \( a_i \) at layer \( l \), computed by the vector message function \( \phi_m^v \). The update functions \( \phi_{\text{un}}^s \) and \( \phi_{\text{un}}^v \) respectively update the scalar and vector embeddings based on the aggregated messages and current state. Likewise, the edge features \( f_{a_i,a_j}^{l+1} \) are updated using \( \phi_{\text{ue}}^s \), which incorporates the vector rejection of embeddings \( \mathbf{v}_{a_i}^l \) and \( \mathbf{v}_{a_j}^l \) along the relative position vector \( \mathbf{r}_{a_ia_j} \), ensuring rotationally invariant geometric representation.

Notably, this approach updates node and edge features during message passing, providing a more comprehensive geometric representation compared to previous studies~\cite{schutt2021equivariant,tholke2022torchmd}.

\paragraph{Residue and Atom Feature Integration.} The residue features \( (\mathbf{e}_{r_i}, \mathbf{z}_{r_i,r_j}) \) and atom features \( (h_{a_i}, \mathbf{v}_{a_i}) \) are integrated and input into an invariant point attention (IPA)~\cite{jumper2021highly} module to capture interactions between residue and atom features:
\begin{equation}
    \mathbf{h}_{r_j} = \text{IPA}(\mathbf{e}_{r_i}, \mathbf{z}_{r_i,r_j}, h_{a_i}, \mathbf{v}_{a_i})
    \label{eq:integration_IPA}
\end{equation}

Furthermore, the IPA module facilitates learning local-to-global relationships within molecular structures by leveraging the interplay between residue and atom features.

\subsection{Diffusion Processes}
\label{sec:diffusion}

Unlike traditional models focused on denoising molecular states, our approach incorporates multi-level molecular features, enhancing accuracy and structural fidelity by leveraging residue- and atom-level information. Let $\mathcal{R}^t = \{ s_j^t, \mathbf{x}_j^t, \mathbf{O}_j^t \}_{j=c+1}^{c+m}$ denote the intermediate states of residue \( j \) at time step \( t \). At \( t = 0 \), the system represents real data, including observed CDR sequences and structures, while at \( t = T \), it corresponds to samples from the prior distribution. Thus, the forward diffusion process adds noise incrementally from \( t = 0 \) to \( t = T \), while the generative process denoises in reverse. Hence, the diffusion processes for amino acid types \( s_j^t \), coordinates \( \mathbf{x}_j^t \), and orientations \( \mathbf{O}_j^t \) are defined as follows:

\paragraph{Multinomial Diffusion for Amino Acid Types.} The forward diffusion process for amino acid types follows a multinomial distribution, formulated as \cite{hoogeboom2021argmax}:
\begin{align} 
q(s_j^t|s_j^{t-1}) = & \mathrm{Multinomial}\Big((1-\beta_t^{\mathrm{type}}) \cdot \mathrm{onehot}(s_j^{t-1}) \nonumber \\
& + \beta_t^{\mathrm{type}} \cdot \frac{1}{20} \cdot \mathbf{1}\Big), \label{eq:d1}
\end{align}
where the onehot function maps an amino acid type into a 20-dimensional one-hot vector, and $\mathbf{1}$ represents an all-one vector. The parameter $\beta_\mathrm{type}^t$ denotes the probability of uniformly resampling an amino acid across 20 types. As $t \to T$, $\beta_\mathrm{type}^t$ approaches 1, driving the distribution closer to uniformity. Thus, the following probability density offers an efficient mechanism to perturb $s_j^t$ at each timestep $t$ during training \cite{hoogeboom2021argmax}:

\begin{align} 
q(s_j^t \mid s_j^0) &= \text{Multinomial} \Big( 
\tilde{\alpha}_{\text{type}}^t \cdot \text{onehot}(s_j^0) \notag \\
&\quad + (1 - \tilde{\alpha}_{\text{type}}^t) \cdot \frac{1}{20} \cdot \mathbf{1} 
\Big), \label{eq:d2}
\end{align}
where $\tilde{\alpha}_{\text{type}}^t = \prod_{\tau=1}^t (1 - \beta_{\text{type}}^\tau).$

The generative diffusion process is defined as:

\begin{equation} 
    p(s_j^{t-1} \mid \mathcal{R}^t, \mathcal{C}) = \mathrm{Multinomial} \left( F(\mathcal{R}^t, \mathcal{C})[j] \right), \label{eq:d3}
\end{equation}
where \( F(\cdot)[j] \) represents a neural network model that takes as input both the structural context (comprising the antigen and antibody framework) and the CDR state from the previous step. This model predicts the probability distribution over the 20 amino acid types for the \( j \)-th residue in the CDR.

Unlike forward diffusion, the generative process relies on the structural context \( C \) and the CDR state from the previous step, which includes positions and orientations. Generative diffusion aims to approximate the posterior \( q(s_j^{t-1} | s_j^t, s_j^0) \), derived from Eq.\ref{eq:d1} and Eq.\ref{eq:d2}, for denoising purposes. Thus, the objective of training the generative diffusion process to predict amino acid types is to minimize the expected Kullback-Leibler (KL) divergence between Eq.\ref{eq:d3} and the posterior distribution. Let \( D_{\text{KL}}^j \) denote the KL divergence for the \( j \)-th term. Then, we define the loss function as:

\begin{equation} 
    L_{\text{type}}^t = \mathbb{E}_{\mathcal{R}^t \sim \mathcal{P}} \left[ 
    \frac{1}{m} \sum_j D_{\text{KL}}^j \right], \label{eq:d4}
\end{equation}
where \( D_{\text{KL}}^j = D_{\text{KL}} \left( q(s_j^{t-1} \mid s_j^t, s_j^0) 
\| p(s_j^{t-1} \mid \mathcal{R}^t, C) \right) \).

\paragraph{\( C_\alpha \) Coordinate Diffusion.} To accurately capture the hierarchical spatial configurations between residues and atoms, our framework models the forward diffusion of normalized \( C_\alpha \) coordinates \( \mathbf{x}_j^t \) as a Gaussian distribution:

\begin{equation}
    q(\mathbf{x}_j^t \mid \mathbf{x}_j^{t-1}) = \mathcal{N}\left( \mathbf{x}_j^t \mid \sqrt{1 - \beta_{\mathrm{pos}}^t} \cdot \mathbf{x}_j^{t-1}, \beta_{\mathrm{pos}}^t \mathbf{I} \right), \label{eq:d5}
\end{equation}

\begin{equation}
    q\left(\mathbf{x}_j^t \mid \mathbf{x}_j^0\right) = \mathcal{N}\left(\mathbf{x}_j^t \mid \sqrt{\bar{\alpha}_\text{pos}^0} \cdot \mathbf{x}_j^0, \left(1 - \bar{\alpha}_\text{pos}^0\right) \mathbf{I}\right),\label{eq:d6}
\end{equation}
where the diffusion rate \(\beta_{\mathrm{pos}}^t\) determines the noise level at each timestep \(t\), increasing from 0 to 1 as the process progresses. The cumulative scaling factor, $\bar{\alpha}_{\mathrm{pos}}^t = \prod_{\tau=1}^t (1 - \beta_{\mathrm{pos}}^\tau).$
By applying the reparameterization trick \cite{ho2020denoising}, the generative diffusion process is formalized as:
\begin{equation}
    p(\mathbf{x}_j^{t-1} \mid \mathcal{R}^t, C) = \mathcal{N}\left(\mathbf{x}_j^{t-1} \mid \mu_p(\mathcal{R}^t, C), \beta_{\mathrm{pos}}^t \mathbf{I}\right),\label{eq:d7}
\end{equation}
where the mean \(\mu_p(\mathcal{R}^t, C)\) is computed as:

% \begin{equation}
%     \mu_p(\mathcal{R}^t, C) = \frac{1}{\sqrt{\alpha_{\mathrm{pos}}^t}} \left( \mathbf{x}_j^t - \frac{\beta_{\mathrm{pos}}^t}{\sqrt{1 - \bar{\alpha}_{\mathrm{pos}}^t}}} G(\mathcal{R}^t, C)[j] \right). \label{eq:d8}
% \end{equation}

\begin{equation}
    \mu_p(\mathcal{R}^t, C) =
    \frac{1}{\sqrt{\alpha_{\mathrm{pos}}^{t}}}
    \left(
        \mathbf{x}_{j}^{t} -
        \frac{\beta_{\mathrm{pos}}^{t}}{\sqrt{1 - \bar{\alpha}_{\mathrm{pos}}^{t}}}
        \, G(\mathcal{R}^{t}, C)[j]
    \right).
    \label{eq:d8}
\end{equation}

where \( G(\cdot)[j] \) is a neural network model that predicts the standard Gaussian noise \( \epsilon_j \sim \mathcal{N}(0, \mathbf{I}) \), which is added to the scaled coordinate of amino acid \( j \), \( \sqrt{\bar{\alpha}_\text{pos}^0} \mathbf{x}_j^0 \), following the reparameterization process described in Eq.\ref{eq:d6}: $\mathbf{x}_j^t = \sqrt{\bar{\alpha}_\text{pos}^0} \mathbf{x}_j^0 + \sqrt{1 - \bar{\alpha}_\text{pos}^0} \epsilon_j.$

The objective function for training the generative process is the expected mean squared error (MSE) between \( G \) and \( \epsilon_j \). This is derived by aligning the distribution \( p \) with the posterior \( q(x_{j}^{t-1} \mid x_{j}^{t}, x_{j}^{0}) \) \cite{ho2020denoising}:

\begin{equation}
L_{\text{pos}}^t = \mathbb{E}\left[ \frac{1}{m} \sum_j \left\| \epsilon_j - G(\mathcal{R}^t, C)[j] \right\|^2 \right]. \label{eq:d9}
\end{equation}

% 强制表格单栏显示
\begin{table*}[h!]
\centering
% 使用 resizebox 将表格缩小适应单栏
\resizebox{\linewidth}{!}{%
\begin{tabular}{@{}l ccc ccc ccc@{}}
\toprule
\textbf{} & \multicolumn{3}{c}{AAR (\%) $\uparrow$} & \multicolumn{3}{c}{IMP (\%) $\uparrow$} & \multicolumn{3}{c}{RMSD (\AA) $\downarrow$} \\
\cmidrule(r){2-4} \cmidrule(r){5-7} \cmidrule(r){8-10}
Method & H1 & H2 & H3 & H1 & H2 & H3 & H1 & H2 & H3 \\
\midrule
RAbD & 22.85 & 25.50 & 22.14 & 43.88 & \textbf{53.50} & 23.25 & 2.261 & 1.641 & \textbf{2.900} \\
DiffAb & \textbf{67.63} & 44.20 & 28.70 & 42.84 & 30.26 & 20.42 & \underline{1.248} & \textbf{1.108} & 3.481 \\
DIFFFORCE & 60.78 & \underline{53.51} & \underline{29.52} & \underline{49.45} & \underline{36.81} & \textbf{30.22} & 1.561 & 1.401 & 3.612 \\
AbMEGD & \underline{66.96} & \textbf{55.40} & \textbf{38.93} & \textbf{55.11} & 24.05 & \underline{23.74} & \textbf{1.169} & \underline{1.130} & \underline{3.419} \\
\bottomrule
\end{tabular}%
}
\caption{Evaluation of the generated antibody CDRs (sequence-structure co-design) by different models. The best result for each metric is highlighted in \textbf{bold}, while the second-best result is indicated with \underline{underlining} for clearer comparison.}
\label{tab:performance_comparison}
\end{table*}

\paragraph{SO(3)-Based Orientation Diffusion.} We propose a diffusion framework based on the \( SO(3) \), extending the method \cite{leach2022denoising} by incorporating multi-level features. The diffusion process on \( SO(3) \) is modeled as an isotropic Gaussian distribution, with dynamics governed by the diffusion parameter \( \beta_{\mathrm{ori}}^t \). This process transitions molecular orientations from well-defined real states at \( t = 0 \) to near-uniform distributions at \( t = T \), simulating the gradual addition of noise \cite{leach2022denoising}:

\begin{equation}
    T_j^t = \mathbf{O}_j^t \cdot \mathrm{ScaleRot}\left( \sqrt{\bar{\alpha}_{\mathrm{ori}}^t}, \mathbf{O}_j^0, 1 - \bar{\alpha}_{\mathrm{ori}}^t \right),
\end{equation}

\begin{equation}
    q(\mathbf{O}_j^t \mid \mathbf{O}_j^0) = \mathcal{IG}_{SO(3)}\left( T_j^t \right), \label{eq:d10}
\end{equation}
\(\mathcal{IG}_{SO(3)}\) represents the isotropic Gaussian distribution on \(SO(3)\), parameterized by a mean rotation and a scalar variance \cite{leach2022denoising,matthies1988normal,nikolayev1997normal}. Then, the ScaleRot operation modifies the rotation matrix by scaling its rotation angle while keeping the rotation axis fixed \cite{gallier2003computing}. \(\bar{\alpha}_{\mathrm{ori}}^t\) = \(\prod_{\tau=1}^t (1 - \beta_{\mathrm{ori}}^\tau)\), where \(\beta_{\mathrm{ori}}^\tau\) represents the variance that increases with the diffusion step \(t\). The conditional distribution used for the orientation generation process is defined as:

\begin{equation}
p(\mathbf{O}_j^{t-1} \mid \mathcal{R}^t, \mathcal{C}) = \mathcal{IG}_{SO(3)}\left( \mathbf{O}_j^{t-1} \cdot H(\mathcal{R}^t, \mathcal{C})[j], \beta_{\mathrm{ori}}^t \right), \label{eq:d11}
\end{equation}
Here, \( H(\cdot)[j] \) represents a neural network designed to denoise orientations, producing the denoised orientation matrix for amino acid \( j \). Training the conditional distribution involves aligning the predicted orientation from \( H(\cdot) \) with the real orientation. Consequently, the training objective is formulated to minimize the expected discrepancy, measured using the inner product between the real and predicted orientation matrices:

\begin{equation}
L_\text{ori}^t = \mathbb{E} \left[ \frac{1}{m} \sum_j \left\| \left( O_j^0 \right)^\top \hat{O}_j^{t-1} - I \right\|_F^2 \right],
\label{eq:d12}
\end{equation}
where \( \hat{\mathbf{O}}_j^{t-1} = H(\cdot)[j] \) represents the predicted orientation matrix for amino acid \( j \).

\paragraph{The Overall Training Objective.} By summing Eq.\ref{eq:d4}, \ref{eq:d9}, and \ref{eq:d12}, and taking the expectation over \( t \), the final training objective function is formulated as:
\begin{equation}
    L = \mathbb{E}_{t \sim \text{Uniform}(1, \dots, T)} \left[ L_\text{type}^t + L_\text{pos}^t + L_\text{ori}^t \right].
\label{eq:overall_loss}
\end{equation}

Training begins by selecting a random time step \( t \) and generating noisy states \( \{s_j^t, \mathbf{x}_j^t, \mathbf{O}_j^t\}_{j=c+1}^{c+m} \sim p \) through the diffusion process, which incorporates noise into the data as described in Eq.\ref{eq:d2}, \ref{eq:d6}, and \ref{eq:d10}. The model parameters are updated by computing the loss based on the perturbed data and applying backpropagation.

\section{Experiments}
We propose AbMEGD, an end-to-end framework that integrates multi-scale equivariant graph diffusion for co-designing antibody sequences and structures. It addresses two key tasks: sequence-structure co-design (Section~\ref{sec:Co-design}) and antibody optimization (Section~\ref{sec:antibody-adopt}). 

\subsection{Sequence-Structure Co-design}
\label{sec:Co-design}

\begin{figure}[hb!]
    \centering
    \includegraphics[width=\linewidth]{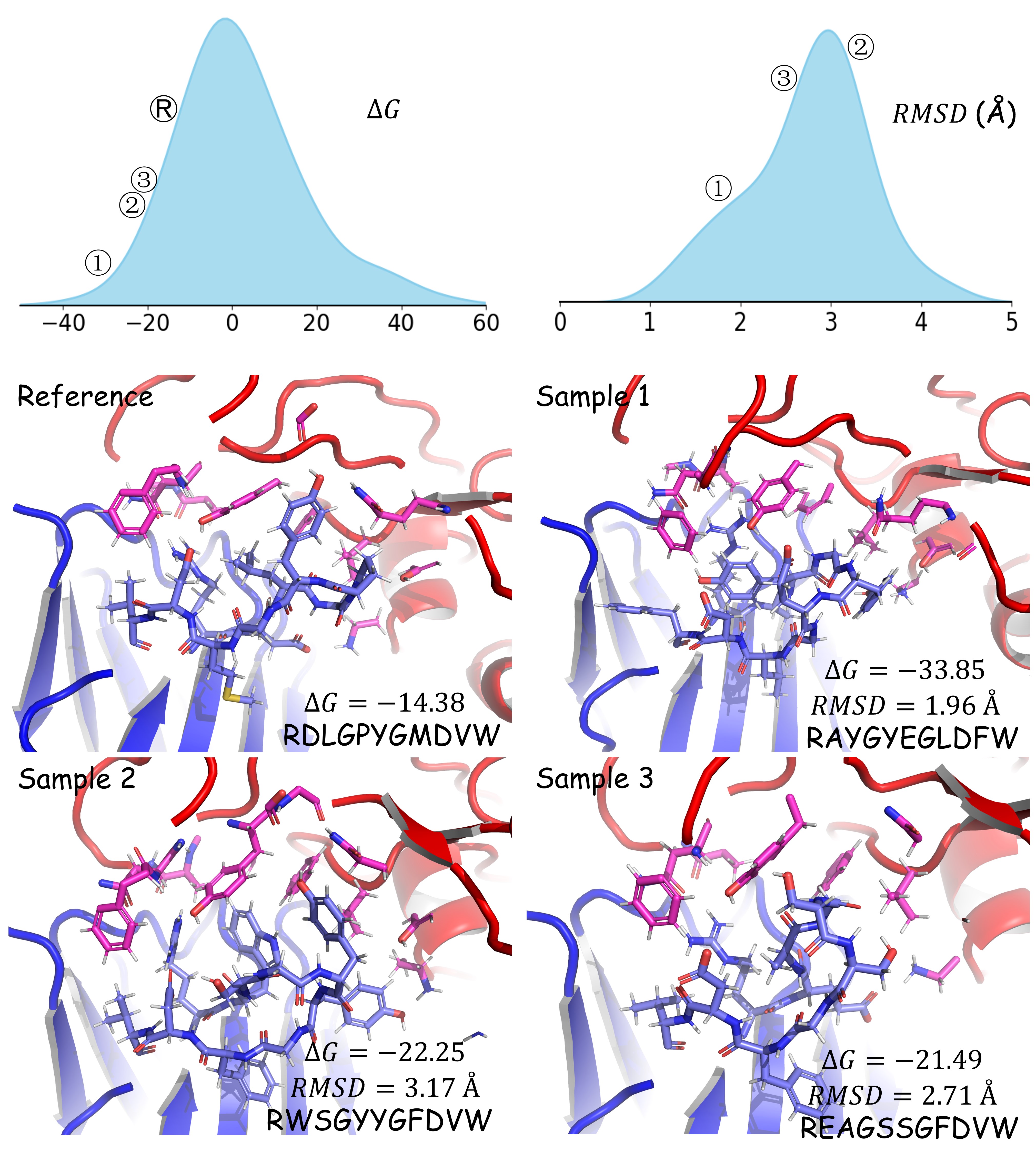} 
    \caption{Examples of CDR-H3 designed using a sequence-structure co-design approach, showcasing the distributions of interaction energy ($\Delta G$) and RMSD. The antigen-antibody complex is derived from PDB entry \href{https://www.rcsb.org/3d-view/7CHF}{7chf}, with the antigen being the SARS-CoV-2 RBD. Sample 1 exhibits better complementarity with the antigen, while Sample 3 displays less effective binding, potentially explaining the variation in their interaction energy ($\Delta G$).
}
    \label{fig:sample}
\end{figure}

We evaluated our model using data from the SAbDab \cite{dunbar2014sabdab}, collected up to September 2024. Antibody-antigen complexes with resolutions worse than 4\AA{} and those binding non-protein antigens were excluded. Antibodies were clustered by 50\% CDR-H3 sequence identity, as in \cite{luo2022antigen}, and five clusters (19 antibody-antigen complexes) formed the test set, including antigens from SARS-CoV-2, MERS, and influenza. The remaining clusters formed the training set.

For evaluation, we removed the CDRs from the test set, generating sequences and structures for the removed regions. We fixed the CDR length to the original value (though other lengths could be explored). Each model generated 100 samples per CDR, with all generated and native structures refined using AMBER99 force field \cite{lindorff2010improved} in OpenMM \cite{eastman2017openmm}  and Rosetta \cite{alford2017rosetta}.

\paragraph{Baseline Models.} To evaluate performance, as described in Section 2.1, we selected three baseline models for comparison. DiffAb \cite{luo2022antigen} represents a diffusion-based generative model specifically designed for antibody structure and sequence modeling. DIFFFORCE \cite{kulyte2024improving} leverages a diffusion framework with force-guided sampling to enhance the quality of generated molecular structures. Additionally, we included the traditional energy-based method RABD \cite{adolf2018rosettaantibodydesign}, which utilizes energy minimization to design antibodies and optimize binding affinities.

\paragraph{Evaluation Metrics.} The models are evaluated using four key metrics: (1) AAR measures how accurately the generated sequences match the native amino acid sequences; (2) RMSD quantifies the structural deviation between the generated and native CDRs by comparing the C$\alpha$ atom coordinates;
and (3) IMP indicates the proportion of designed antibodies that achieve improved binding energy ($\Delta G$) compared to the baseline, calculated using the InterfaceAnalyzer module in Rosetta.

Table~\ref{tab:performance_comparison} compares four antibody design models across three metrics. AbMEGD excels overall, leading in AAR for H2 and H3, with competitive results in H1. It achieves the lowest RMSD in H1, indicating high structural accuracy, and performs well in H2 and H3 RMSD. While DIFFFORCE slightly outperforms AbMEGD in IMP for H2 and H3, AbMEGD leads in H1, balancing binding energy improvement with sequence recovery. DIFFFORCE demonstrates its strength in affinity optimization, particularly in H3, while DiffAb performs well in AAR and achieves the lowest RMSD in H2 but lags in IMP. RAbD shows moderate IMP in H2 but low AAR, indicating poor sequence recovery, with favorable RMSD only in H3 due to reliance on the Rosetta energy function. In contrast, AbMEGD achieves robust energy improvements without Rosetta supervision. Building on DiffAb, which is widely recognized as a leading model in antibody design, AbMEGD significantly improves performance in the critical CDR-H3 region, achieving a 10.13\% increase in AAR (from 28.70\% to 38.93\%), a 3.32\% boost in IMP (from 20.42\% to 23.74\%), and a 0.062 \AA\ reduction in RMSD (from 3.481 \AA\ to 3.419 \AA). These results highlight AbMEGD’s capability to refine sequence-structure co-design and affinity optimization, surpassing current models. Figure~\ref{fig:sample} illustrates three designed examples of CDR-H3 interacting with the SARS-CoV-2 RBD. Among these, Sample 1 exhibits the strongest binding affinity, as reflected by its lowest binding energy, and demonstrates superior structural complementarity with the antigen. In contrast, Sample 3 shows a higher binding energy compared to the reference, and its CDR shape appears less compatible with the antigen, indicating weaker interactions.
\begin{figure}[hb!]
    \centering
    \includegraphics[width=\linewidth]{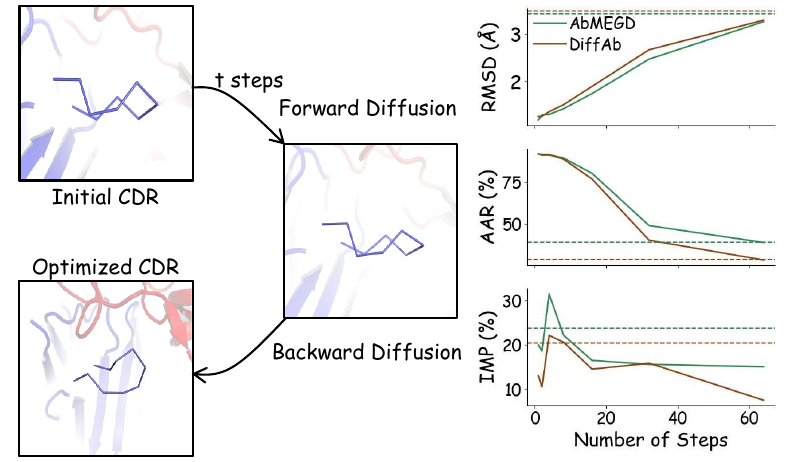} 
    \caption{(a) The antibody fenoptimization algorithm applies a forward diffusion process to perturb the initial CDR over $t$ steps, followed by a backward diffusion process to refine it into the optimized CDR. (b) The IMP, RMSD, and SeqID metrics of the CDRs were evaluated across varying optimization steps. The dashed lines indicate the performance of CDRs generated through de novo design. At $t = 4$, the optimized CDRs achieve an IMP score comparable to that of the de novo designs while retaining a structural similarity to the original CDR.}
\label{fig:diffusion_steps}
\end{figure}

\subsection{Antibody Optimization}
\label{sec:antibody-adopt}

\begin{table}[htbp!]
\centering
% 使用 resizebox 将表格缩小适应单栏
\resizebox{\linewidth}{!}{%
\begin{tabular}{@{}c cc cc cc@{}}
\toprule
\textbf{} & \multicolumn{2}{c}{\textbf{RMSD (\AA) $\downarrow$}} & \multicolumn{2}{c}{\textbf{AAR (\%) $\uparrow$}} & \multicolumn{2}{c}{\textbf{IMP (\%) $\uparrow$}} \\
\cmidrule(r){2-3} \cmidrule(r){4-5} \cmidrule(r){6-7}
\textbf{Steps}& \textbf{AbMEGD} & \textbf{DiffAb} & \textbf{AbMEGD} & \textbf{DiffAb} & \textbf{AbMEGD} & \textbf{DiffAb} \\
\midrule
1  & 1.166 & 1.239 & 91.63 & 91.60 & 20.00 & 13.04 \\
2  & 1.274 & 1.244 & 90.97 & 91.23 & 18.64 & 10.53 \\
4  & 1.284 & 1.342 & 91.18 & 91.19 & 31.48 & 22.18 \\
8  & 1.404 & 1.490 & 89.27 & 88.58 & 22.24 & 20.66 \\
16 & 1.728 & 1.880 & 80.09 & 76.70 & 16.53 & 14.56 \\
32 & 2.458 & 2.657 & 48.97 & 40.24 & 15.64 & 15.89 \\
64 & 3.253 & 3.286 & 38.91 & 28.44 & 15.11 &  7.58 \\
T  & 3.419 & 3.481 & 38.93 & 28.70 & 23.74 & 20.42 \\
\bottomrule
\end{tabular}%
}
\caption{Evaluation of optimized CDR-H3s with different optimization steps. This table presents metrics including RMSD (Å), AAR (\%), and IMP (\%) to assess the performance of the optimization approach. Unlike redesigning methods, the optimization process improves binding energy while retaining structural similarity of the optimized CDR to the original one. Results for RMSD, AAR, and IMP are reported for AbMEGD and DiffAb. The accompanying line plot in Figure~\ref{fig:diffusion_steps}b further illustrates these results.}
\label{tab:adopt_table}
\end{table}

Our model is designed to optimize existing antibodies, providing an important application in the pharmaceutical domain. The optimization process begins by perturbing the CDR sequence and structure for $t$ steps using the forward diffusion process. Afterward, starting from the $(T - t)$-th step (with $t$ steps remaining) of the generative diffusion process, the sequences are denoised to produce a set of optimized antibodies. This procedure is illustrated in Figure~\ref{fig:diffusion_steps}a.

We focus specifically on optimizing the CDR-H3 antibody region in the test set with varying $t$ values. For each antibody and $t$, the CDR is independently perturbed 100 times, yielding 100 optimized CDRs distinct from the original CDR. We evaluate the performance of our model and DiffAb by reporting the percentage of antibodies with IMP, RMSD, and AAR of the optimized CDRs relative to the original antibody. Additionally, we compare the optimized antibodies with \textit{de novo} antibodies ($t = T = 100$) introduced in Section~\ref{sec:Co-design}.

As depicted in Table~\ref{tab:adopt_table} and Figure~\ref{fig:diffusion_steps}b , our optimization method produces antibodies with superior binding energy (IMP) as measured by the Rosetta energy function. At $t = 4$, the optimized CDRs reach an IMP score comparable to that of de novo antibodies, while maintaining significantly higher structural similarity to the original CDRs. The RMSD and AAR results further demonstrate that our model achieves a better trade-off between binding energy improvement and structural conservation compared to DiffAb, making it more suitable for practical applications.

\section{Conclusion}
In this study, we propose the AbMEGD, an advanced diffusion-based generative model addressing key challenges in antibody design. By integrating residue- and atomic-level features into a unified framework, AbMEGD achieves high performance in sequence-structure co-design and antibody optimization. The experimental results confirm its ability to balance structural conservation and enhance functionality during antibody refinement.

\section*{Ethical Statement}

There are no ethical issues.

\section*{Acknowledgments}

This work was supported in part by the Natural Science Foundation of China (No.62476203), and the Guangdong Provincial Natural Science Foundation General Project (No.2025A1515012155). Key Program of Hubei Natural Science Foundation Traditional Chinese Medicine Innovation and Development Joint Fund‌ (2025AFD470).

% \clearpage
% \clearpage

%% The file named.bst is a bibliography style file for BibTeX 0.99c
\bibliographystyle{named}

\end{document}